# Artificial Intelligence, conceptual metaphors and conceptual engineering: Are AI-based framings of human behaviour and cognition successful?

WARMHOLD JAN THOMAS MOLLEMA[1] & THOMAS WACHTER[2]

**Abstract**

Understanding human behaviour, neuroscience and psychology using concepts from the domain of AI is increasing in popularity. Given the massive integration of AI technologies into our daily lives, AI-related concepts are being used to compare AI systems with human behaviour, brain functions, and cognitive abilities like language acquisition. But scientists and philosophers are also increasingly tempted to take the AI-framing of the human conceptual domain as a literal one. This paper investigates the epistemic and practical success of these 'AI-framings': *What does it mean to apply the conceptual constellation of AI to the human conceptual domain?* We consider and compare two possible answers: either these examples are *conceptual metaphors*, or they are attempts at *conceptual engineering*. Firstly, we argue that when viewed as conceptual metaphors, the AI-framed descriptions risk committing the "map-territory fallacy". Secondly, we argue the comparisons also contain a misleading 'double metaphor' because of the metaphorical connection between human psychology and computation at the conceptual foundation of computation. But we also argue that there is a possible semantic catch to the AI-framing, which is captured by the conceptual engineering view. This is that the AI-framings point towards avenues for forms of conceptual engineering. If the challenges of conceptual ethics and reductionism are overcome, some AI-framings might enrich our epistemic and practical lives. So, at its worst - as implicit conceptual metaphor - the AI-framing leads us completely astray; at its best, it prompts us to reflect anew on how the boundaries of our current concepts serve us and how they could be improved.

## 1. The conceptual human-AI-mess

Angie Wang (2023) sketches a story of a mother that perceives parallels between her toddler's process of learning language and the 'stochastic parroting' of LLMs (Bender et al., 2021). When she witnesses her son stringing together vowels and calling many different objects 'cat', she directly employs AI-language to make sense of his behaviour. She asks: "Aren't we after all just a wetware neural network, a complex electrochemical machine?". When she corrects him calling everything 'cat', she asks: "So is this supervised learning?"; "Is my Toddler a Stochastic Parrot?". Wang's mother-protagonist, at a moment of despair, comes to the following: "ChatGPT is a stochastic parrot and so are we." Like OpenAI's CEO Sam Altman (2022) once tweeted: "i am a stochastic parrot and so r u". Since most of the dominant debate revolves around whether, theoretically and ethically, AI can and should be described with concepts from the human and animal conceptual domains, less attention has been devoted to the reverse question: whether the human domain can and should be understood in terms of AI concepts.

[1] Vrije University Amsterdam (Department of Philosophy) Amsterdam, The Netherlands. wjt.mollema@gmail.com

[2] National Centre for Artificial Intelligence (CENIA), Santiago, Chile.

AI-based explanations of human behaviour and cognition are gaining credit. On the one hand, given the increasing integration of AI in daily life, AI imagery has become more salient or *vorhanden* for laypeople to describe human behaviour and cognition. On the other hand, professional researchers, make serious philosophical claims. For instance, Christopher Summerfield states that "Turing's question [Can machines think?] has a trivial answer. Machines that think already exist. They are called *brains* and there are many trillion of them on this planet." (Summerfield, 2023, p. 5). Similarly, Blaise Agüera y Arcas, while not claiming "That the brain is a Transformer", claims like: "the Turing completeness of the Transformer model, the breadth of intellectual moves it can readily learn, and the ability to chain those moves using short-term memory are the key prerequisites for general intelligence" (Agüera y Arcas, 2025, ch. 8)., This is an example of framing human concepts in terms of AI concepts. With respect to cognitive science for example, as Valerio Capraro (2026) points out: "Terms such as "prediction", "pattern completion", "generation", "prompting", "hallucination", "training data", and "nexttoken prediction" are not merely technical descriptors of artificial systems: they can become culturally available descriptors of human thought, creativity, memory, and explanation". This tendency, so we argue, can be understood as a new iteration of the computer-metaphor in cognitive science and psychology (Baria & Cross, 2021; Brette, 2022; Piccinini, 2009; Shagrir, 2006), but shouldn't plainly be dismissed. Rather, it's important to come to a better understanding of what it means methodologically to apply these AI-related concepts to speak about human cognition and behaviour.[3]

[3] There is already some critical work on the dangers of framing cognitive science in terms of AI. Authors like Floridi and Nobre (2024) have recently discussed how the use of AI-related concepts in cognitive science can obscure important aspects of our cognition. From an ethical and practical point of view, this issue calls for a careful consideration of the social implications that this could bring. Mitchell (2024a) notes that the use of the metaphors used to describe AI systems affects the way we interact with them, how much we trust them and also the different ways we

So, the niche this paper occupies is that its driving question is exactly the reverse of the dominant discussion: *what does it mean to apply the conceptual constellation of AI to the human conceptual domain?* We are interested in what epistemic guidance these concepts provide and how these scientific and lay attempts at understanding human behaviour, psychology and neuroscience in terms of AI can be understood. This line of investigation is similar to, but more encompassing than the one explored by Valerio Capraro (2026) who coined the term "LLMorphism", a variant of "mechanomorphism", to express "the biased belief that human cognition works like a large language model".

Since these are big questions, and papers are short, we devote ourselves to discussing only two interrelated answers, which we approach comparatively: can these AI-framings be conceptualized as *conceptual metaphors*, or they are attempts at or cues for *conceptual engineering*? Arguably, both views are on the same spectrum, but as section 3 will make clear, conceptual engineering is a practice with the explicit goal of revising concepts, producing new concepts or abandoning concepts, and a conceptual metaphor is a structural, non-literal description of some $X$ being $Y$.

Before developing these questions, we want to flag some assumptions that co-determine the scope of our discussion. In the philosophy of mind and technology, proponents of some variants of 'computationalism' (whether construed as functionalism about mental states or not) think that thinking of the mind as working like a computer is literally apt. There are also other forms of literalism about the brain-computer description. And one can believe that the brain computes without being a computationalist (although not vice versa).[4] While we will not argue against this position, we assume it is fallacious. This is important to flag, because due to this we deliberately leave an important position out of the discussion. One could namely also opt for some sort of computational literalism about AI-based framings of human behaviour and cognition. Instead, we compare and discuss the other non-literal approaches - the conceptual metaphor and conceptual engineering views. So, while we do not assume metaphorical or revisionary are the only ways to go, we think they deserve our attention here decoupled from cases for literalism. But note that the epistemic success of neither of these approaches depends on the aptness of computationalism as conceptual framework; qua perspective, they are not rooted in a specific

---

regulate them. In this sense, the philosophical questions treated in this paper are closely tied with contemporary discussions on AI ethics. Still, we have not encountered scholarly work comparing the conceptual metaphor and conceptual engineering views on this topic.

[4] We are indebted to two reviewers for prompting us to clarify these assumptions and helping us disentangle literalism from computationalism.

position in the philosophy of mind. Another assumption to declare is that we believe the varieties of computational structures of the AI domain are rich and varied but not generally representative for the computational domain as a whole. This leads us to view AI systems, and LLMs in particular, as not being exemplary for computational thinking, but as a rather peculiar region of it.

To answer our main question, we proceed as follows. Section 2 historicizes the application of the metaphors of 'machine' and 'computer' to the human domain. This provides a valuable historical background for making sense of the contemporary AI-based descriptions of the human domain. Against this backdrop we define two possibilities for viewing the deployment of AI concepts: the conceptual metaphor view (*CM*) and conceptual engineering view (*CE*). Section 3 discusses standard accounts of conceptual metaphor theory and conceptual engineering to deepen our understanding of both. Section 4 covers our main argument regarding *CM*'s fit to the stochastic parrot case. We argue, firstly (4.1) that these framings risks committing the "map-territory fallacy". Secondly, at the level of the conceptual foundations of computation, (4.2) we show this computational style of explanation is a 'double metaphor' that is epistemically misleading if not explicitly recognized, because of the metaphorical connection between human psychology at the root of computation. To complete the comparison, section 5 deals with *CE*: at least three ways of conceptual engineering in light of the standard account can be envisioned in this context, as well as two challenges that are especially significant for the AI-context.

## 2. The machine conception of organism, AI and metaphor

Few notions in biology have been so pervasive as the machine conception of the organism (MCO). Formulated by Descartes in the seventeenth century, the MCO is based on the metaphorical redescription of organisms as machines to understand them (Nicholson, 2013). For instance, in 1665 the Danish anatomist Nicolaus Steno boldly argued that if we want to understand how and what the brain does, we should take it apart and, with that, view it as a machine (Cobb, 2020). From there, the history of theorising organisms became entangled with technological development, constantly comparing ourselves, metaphorically, with the most advanced technology of the time (Birhane, 2021).[5] This section discusses the origins of the MCO's contemporary AI-manifestation and culminates in the stipulation of two ways to understand: as conceptual metaphor or conceptual engineering.

[5] In the seventeenth century, it referred to clockworks—intricately calibrated parts working together as a single, integrated unit. By the eighteenth century, it described steam engines, which consumed energy through combustion to perform work and generate heat. In the nineteenth century, the term extended to chemical factories, where machines coordinated and managed complex networks of interconnected chemical reactions (Nicholson, 2013).

Since the advent of the computer in the 1950s, the study of brain and cognition has been dominated by concepts from information theory and computer science (Floridi & Nobre, 2024). As Freek Bomhof (2026) has written, machinic metaphors have been continuously introduced, from steam engines through electromechanical machines[6] towards computers: "I have to process this", or "let me update you". People have to "reboot" sometimes." In science, concepts like "information processing", "coding", "computing", and "algorithm" frame the brain and mind as computational information processing systems (Brette, 2022; Colombo & Piccinini, 2024; Floridi & Nobre, 2024). This new version of the MCO has been instrumental to the development of cognitive science, psychology and AI as a science (Felin & Holweg, 2024).[7]

The scientific study of AI is strongly influenced by MCO, as it approaches intelligence as an inherently computable phenomenon. For example, according to the field's founders (convening at the Dartmouth Conference in 1956), their goal was to "proceed on the basis of the conjecture that every aspect of learning or any other feature of intelligence can in principle be so precisely described that a machine can be made to simulate it" (McCarthy et al., 2006). Moreover, McCulloch and Pitts' first mathematical model of the neuron – foundational to modern AI –, was directly inspired by the *analogy* of the brain as a computer in which neurons were modelled like logic gates (Chirimuuta, 2021).[8] Despite criticism stressing the metaphorical relation of the conceptual connection of brains and AI, this conceptual framework remains highly regarded in cognitive science. Some argue that it provides valuable conceptual and formal tools for theory development and enables meticulous assessment of computational feasibility (Van Rooij et al., 2024). As Margaret Boden (1988) puts it: "Computers as such are, in principle, less crucial for cognitive science than *computational concepts* are" (our emphasis). More recently, Blaise Agüera y Arcas (2025) has argued that, while acknowledging that "there are meaningful differences between biological computing and the kind of digital computing done by the ENIAC or your smartphone", says that "It's not a metaphor to call DNA a "program"—that is literally the case."

---

[6] Bomhof notes the following examples for the steam engine: going "full steam ahead"; we "run out of steam"; sometimes we "need to "blow off steam" in order to prevent that we "go off the rails"." For electromechanical machines, his examples are: "we are on the same wavelength"; being "off the hook"; "ideas are "sparked""; and our brains sometimes "short-circuit".

[7]For example, in the influential book that initiated the field of Cognitive Psychology, Ulric Neisser states: "The task of a psychologist trying to understand human cognition is analogous to that of a man trying to discover how a *computer has been programmed*" (Neisser, 1967, p. 6, emphasis added).

[8] It is important to mention that they believed that, under certain conditions, the brain was literally a computer (Chirimuuta, 2021). As noted in the introduction, there are multiple ways to interpret this 'literalism'.

The practice of stressing similarities between organisms and computational machines comes with the temptation to construe the framing of organismic concepts as computational as a *metaphysical* one. But, we think that, whether or not the framing is intended literally, it should be resisted, as it's at best a contingent epistemological perspective; but a successful one at that. The logico-grammatical conceptual organisation of the computational is one 'we' – those bombarded with computers and AI – come to find attractive, but for which alternatives are possible. In short, the AI-manifestation of MCO ('animal cognition is conceived of as digital machinery') is highly relevant to the literal and metaphorical positions. Let's call this 'AI-framing'.

To further this discussion, we extend it by specifying and thinking through two new ways to approach the conceptual exchange between the cognitive domain and the AI-domain. First, and negatively, today's understanding of human behaviour and psychology in terms of AI could be a *conceptual metaphor*. One of our conclusions is that leaving this implicit in the current discourse has significant downsides. Appeal to metaphor namely emphasises certain convergences at the cost of suppressing divergences thereby being necessarily inexhaustive for the target domain. Second, and positively, AI-framing of cognition and behaviour could be the start of applying the methodology of *conceptual engineering:* the pragmatic, moral and epistemic forms of maintenance and improvement to our conceptual repertoire in the context of AI.

Let's unpack both *CM* and *CE*.

Understanding AI-framings as *conceptual metaphors* is to view them as systematic projections of one conceptual domain onto another for highlighting similarities between both domains (Kövecses, 2017). Put differently, the conceptual domain of the AI is structurally organised and deployed to analogically approach the domains of life, mind and humanity. Call this:

> *CM*: Using AI-concepts to explain human behaviour, cognition and psychology should be understood as *the projection of the former conceptual domain onto the latter domains, which highlights similarities, while suppressing differences between them*. This is an epistemological *comparison* because it structures the way in which we can know the target domain.

Alternatively, these attempts can be viewed as *conceptual engineering* – designing or otherwise changing concepts in terms of their content, usage, inferential relations, or intended reference (Cappelen, 2018; Hopster & Löhr, 2023). That is: understanding concepts from the human conceptual domain in terms of AI is the crafting, amelioration or elimination of concepts belonging to the conceptual constellation of the mind. Call this:

> *CE*: Using AI-concepts to explain human behaviour, cognition and psychology should be understood as shifting conceptual boundaries for *extending, eliminating or refining existing concepts to better fit reality* (in some way). Extant concepts are altered, replaced or eliminated because the conceptual resources of the computational and AI provide better models for these domains.

Returning to the 'stochastic toddler' and similar examples, we observe an initial tension between *CM* and *CE*. First, on *CM*, there is a structurally recurring comparison between the 'AI' qua conceptual family and with the conceptual families of BEHAVIOUR and COGNITION.[9] The 'stochastic toddler' seems like a symmetrical reversal of the "'LLM as individual mind' metaphor" (Mitchell, 2024a). *Prima facie* this is a moderately/strongly epistemically successful conceptual metaphor (see section 3) as it intuitively 'seems to work'. We will critically examine the extent to which it *can* work in sections 4.1.-4.2. Conversely, under *CE*, talk of toddlers as being stochastic parrots (etc.) can be seen as stretching and pushing conceptual boundaries from the human domain to AI towards a new stable state. If this talk is interpreted as providing cues for the conceptual re-engineering of human domains via AI concepts, we could change our language-games accordingly thereby shifting the inferential content of concepts and their interconnections. The effects would be bidirectional, e.g. affecting, on the AI-side, 'reinforcement learning', 'stochastic language model', etc., and on the cognitive, psychological and behavioural end concepts like 'learning', 'language production', or 'reasoning'.

In this paper's remainder, we thoroughly set up both views (section 3) and compare their feasibility in terms of (a) explanatory applicability to the human-AI nexus and (b) prospects for practical application (sections 4 and 5).

Hint with respect to our takeaways: We think that cases like 'stochastic toddler' are currently better explained as *CM* rather than by *CE* because of their *intuitive* rather than *methodological* use in practice. This conceptual metaphor comes with limitations that should be more widely recognized (see section 4). Conceptual metaphors are insightful models only to a certain, often aesthetically enabled sense. In contrast to a specific representational device that has been engineered for a maximum of representational adequacy (the ultimate *CE* output), with *CM* we need to be more epistemically cautious. While there is of course no perfect model (so that's not a standard we can apply to *CM*), it is clear that AI provides not the only, and probably not the best model for human behaviour and cognition. Currently, we are socioculturally initiated into the usefulness of talk, models, and explanations based on AI to make sense of human domains. This practice has proliferated to the point that it has become intuitive for many language-users. Differently stated, this particular use of concepts is in the process of *fossilisation*, becoming a certainty for many of us (Wittgenstein, 1975, §474). But remaining critical of this process means acknowledging that any language-game can in principle be played differently, while still satisfying its reason for being played: in our case the scientific and lay understandings of the human domain. It could be argued that where it is a step back the AI-framing should be resisted, and that where it leads to new insights, promising novel theories, etc., we should embrace it as a conceptual engineering project. The discussions in sections 4-5 aim to make this question clearer, by examining how the *CM*-construal of AI-framings of the human domain come with epistemic instabilities of and the *CE*-view faces specific challenges.

[9] Following (Lakoff & Johnson, 2003), strings that indicate (components of) conceptual metaphors are written in uppercase, connected by lowercase verbs.

## 3. Conceptual metaphor theory and conceptual engineering

This section provides the theoretical background necessary for understanding *CM* (conceptual metaphor theory, see section 3.1.) and *CE* (conceptual engineering as revisionary philosophical methodology, see section 3.2.).

### 3.1. Conceptual metaphors

Max Black's (1955) 'interaction theory of metaphor' revived metaphor theory and conceptualised a metaphor as having a *principal* subject ('target domain') and a *subsidiary* subject ('source domain'). In the case of the metaphor 'that man is a wolf', the principal subject is 'that man' and the subsidiary subject is 'a wolf'. When the metaphor is evoked, it "selects" or "filters" aspects from a target domain by projecting the wolf-"system of commonplaces" onto the source domain (man) (Black, 1955, pp. 73–75).[10] Every metaphor is "the tip of a submerged model" (Black, 1977, p. 445): a picturing of something from a certain perspective, with explanatorily valuable expressive power. A successful metaphor discloses the target domain's "intelligible structure" (Miller, 1979). So a metaphor's provided model is an analogical organisation of two conceptual domains for stressing certain similarities, while suppressing certain differences.[11]

What is a *conceptual* metaphor then? Conceptual metaphor theory, developed by George Lakoff and Mark Johnson (1981; 2003) builds upon Black's conceptualization. A conceptual metaphor is a *set of correspondences* between source domain and target domain (Kövecses, 2017; Lakoff, 1993; Lakoff & Johnson, 2003) that enables a structural understanding of "coherent organization[s] of human experience" (Kövecses, 2008). Common examples are 'TIME is MONEY', 'LOVE is a JOURNEY', 'ARGUMENT is WAR', 'UP is BETTER' or 'THEORY is a BUILDING', and there are many ways in which the resemblance between the source and target domain is spelt out in everyday parlance. E.g., many variants of expressions like 'I'm *short* on time', 'she *attacked* the premise', or 'the neuron is *fundamental* to neuroscience' are commonly used. Conceptual metaphors, therefore, make two domains 'emphasise and resonate' (Black, 1977, pp. 439-440). Furthermore, conceptual metaphors aren't arbitrary: they "are grounded in systematic correlations within our experience" (Lakoff & Johnson, 2003, p. 61). They go over and above individual metaphors, because they express a descriptive invariance that can be constructed between two domains. It is a regular pattern of similarities running through our everyday parlance and thought.

[10] Of course, Black's view was not without critics. It is out of scope however to engage in the debate on this matter.

[11] This invitation to take a certain perspective is more than merely aesthetic: a "distinctive mode of achieving insight", enabling us "to *see new connections*" (Black, 1962, pp. 223; 236-237).

A metaphor's perspective can be more or less epistemically successful. Metaphors also play scientific roles (Taylor & Dewsbury, 2018; Rivadulla, 2006); think of the atom, electricity as a current, the heart as a pump, etc.. To explain how metaphorical projections can be inescapably flawed, while still being epistemically successful, Nikola Kompa (2021, p. 41) has proposed criteria to measure the epistemic success of metaphors with:

whether the entities in the metaphor exist;
if inferential patterns are present between domains;
the extent of the disanalogies between the domains;
the theoretical merits of the metaphor's perspective;
the extent to which necessary properties of the target domain are obscured;
whether it engenders new metaphors.

Intuitively, whether or not these criteria are met, a conceptual metaphor at least discloses knowledge of some target domain in a structurally successful way: it encompasses a set of correspondences, while a 'normal' metaphor is tailored to a specific context.[12] However, conceptual metaphors also have limits. We may talk about love *as if* it's structured like a journey, but love also has dispositional, microbiological and psychological aspects, unshared with journeys. In short, "we prefer to exploit *contingent* properties, not *defining* properties, metaphorically" (Kompa, 2021, p. 34). Metaphor is "a seeing-as which occurs only while I am actually concerning myself with the picture as the object represented" (Wittgenstein, 2009, II, §199); a framing of one concept by another that depends on an agent implementing said perspective of the target concept.

With this exposition of conceptual metaphors in mind, we turn to a discussion of conceptual engineering.

[12] Calling a man a wolf clearly works differently than thinking time analogously to money. For example, there are numerous expressions that are part of TIME is MONEY, such as 'I'm short on time' or 'time is scarce', etc.

### 3.2. Conceptual engineering

Conceptual engineering is a longstanding philosophical practice (Löhr, 2024): proposing new concepts, disposing of old ones, or redrawing the boundaries of existing concepts. Changing our networks of concepts not only concerns tailoring concepts to the 'joints of nature', but also requires ethical reflection, i.e., the evaluation of concepts. Conceptual ethics is the normative reflection on whether the concepts we have, are the concepts we *should* have, if the boundaries they lay down for us to go on with about our lives suffice for our needs (Queloz, 2021). And it is about how the concepts we use now are related to "us" as a social group, distinct from other social groups (Mollema, 2024a; Queloz & Cueni, 2021). Some hold that we have come to have *these* concepts because of underlying needs to use them thus and so and that we should continue to tailor our talk and thought to those needs (Queloz, 2025). While it is helpful to distinguish engineering and ethics, the conceptual engineer often fulfils both roles (Löhr, 2024), and for matters of simplicity, we treat the methodology of conceptual engineering as relating to both logical-grammar and ethics.

Also, there are many different accounts of conceptual engineering available. None is uncontroversial. For our present purposes, we have synthesized a quasi-standard picture of conceptual engineering that assumes the following:

- conceptual engineering is about adding, revising and eliminating concepts in one's repertoire in some way;
- concepts have some sort of content (whether internal or external doesn't matter now) which we can loosely talk about in terms of *intension* and *extension* or in terms of *inferential connections* between concepts;
- conceptual engineering of terms and their intensions, extensions or inferential relationships is possible, from within existing linguistic practices;
- conceptual engineering impacts how we think, talk and know mediated by conceptual changes;
- concepts have inferential as well as epistemic roles in linguistic practice and implicitly co-structure non-linguistic practices;
- conceptual pluralism is possible (we can devise different conceptualizations of the same extension).

Again, here we do not aim to argue for a best way to engineer concepts. Rather, we want to flesh out what it would look like to do so in the context of the AI-framings of human cognition and behaviour. This 'toy model' of the methodology serves that specific purpose.

Defining concepts that figure in everyday life is often hard, impossible, or misguidedly futile. As Wittgenstein (2009, §569) says: "concepts are instruments" that enable and determine how we can go about and while some may fit the same purpose, others take "more time and trouble than we can afford." The contemporary debate on conceptual engineering therefore isn't preoccupied with achieving so-called 'conceptual hygiene', but rather with how conceptual engineering ties in with forms of social change (Hopster et al., 2023). For example, Hopster & Löhr (2023, p. 8) argue that conceptual disruptions are the roots of conceptual engineering. An unsettlement of how a concept is collectively understood is what re-engineering our conceptual network addresses. In the context of AI that concerns us, conceptual engineering for example responds to the wind of societal disturbances that AI brought along.

Simplified, these disturbances come in three forms:

⇒ *gaps*: the conceptual tools are lacking for subsuming new phenomena (Veluwenkamp et al., 2024). For example, a natural anomaly is observed, like a new type of particle, or unforeseen circumstances occur, like the invention of a new technology.

⇒ *mismatches:* a conceptual delineation is deemed unfit for its current purpose(s). For example, using the concepts of 'soul' and 'mind' to account for human cognition and psychology was once suitable, but ended up in mismatch with mainstream scientific perspectives. What Hopster & Löhr (2023, p. 10) call 'misalignment' is a specific case of a mismatch, where "a given concept that is entrenched in a joint conceptual scheme is insufficiently aligned with the overall goals (prudential, moral, etc.) of the agents who deploy it".

⇒ *improvements:* a new conceptualization is proposed to supersede the status quo. Such a proposition can serve to address a 'conceptual overlap' (Hopster & Löhr, 2023). Think of the semantic recharge of the concept of 'gender' in the 1960s and its subsequent differentiations that allowed for novel granular expressions of identity.

Conceptual gaps, mismatches or improvements arise via technological, socio-cultural or psycho-biological change. These changes can be gradual and general, as well as *ad hoc,* because historically philosophers, scientists and politicians have made a wealth of different changes to our conceptual network. In the context that concerns us, AI-framings might provide the start of a conceptual engineering project by being targeted at gaps or mismatches or be framed as improvements.

To complete the introduction of conceptual engineering, we map gaps, mismatches or improvements to how (1) we can change existing concepts for the better; (2) introduce new concepts; or (3) eliminate existing concepts.

(1) Revising existing concepts for the better is termed 'conceptual amelioration' by Haslanger (2020) and 'adaptation' by Hopster & Löhr (2023). As Haslanger (2020, p. 230) says: "we should seek not only to elucidate the concepts we have but aim to improve them in light of our legitimate purposes". Let's take Haslanger's approach as a springboard for our present illustration. Thereby we take the key points from her account and frame them slightly broader, informed by our general picture of conceptual engineering. These improvements come in *epistemic* as well as *semantic* forms that respectively concern changing concepts' knowledge structures and acquisition, and changing a concept's partitioning of logical space. Epistemic amelioration targets how and what we can know about our representational concepts, which can have alethic consequences, whereas semantic amelioration adjusts conceptual intensions or inferential entry clauses, which can have epistemic consequences. We can epistemically (i) refine a concept to gain better knowledge of what it represents to us and (ii) improve our experiential access to the concept's informational content (Haslanger, 2020, p. 242). Semantically, we can (iii) socially reconstruct concepts so that concept-users can deploy them; (iv) pragmatically rearrange the conditions of coordination of a concept (when, where and how it is appropriate to use it); and (v) change the ethical significance of a concept's logical space. In short, amelioration/adaptation can alternatively be described as forms of rational preservation of existing concepts. Sometimes, these are multiple different motivations for making the (same type of) conceptual change.

Consider the concept 'spider' as an illustration of these five cases. We might want to narrowly define SPIDER to better understand what it represents to us and avoid the confusion with, e.g., Japanese spider crabs. The scientific denomination *Araneae* does this by epistemically refining SPIDER - also called 'explication' in other contexts. However, users of the word 'spider' may have never encountered one. In that case we could improve their representational access to what SPIDER is about by showing them one.[13] We could also socially rearrange SPIDER's conceptual delineation to exclude the property of having eight legs. This would be apt were a new species of nine-legged spiders to be discovered. Now imagine a more scientifically stringent society; in that case, the conditions of coordination of the word 'spider' could be changed: we can determine that it denotes the *'order' abstraction* and is no longer fit to be used to describe '*species' abstractions* like *Tegenaria domestica* or *Poecilotheria metallica*. Finally, 'spider' might change in ethical significance and come to be seen or used as a morally harmful one, for example because of collective associations with a terrorist group called 'The Spiders'.

(2) Fabricating new concepts – called 'lexical expansion' (Cappelen, 2020) or 'introduction' (Hopster & Löhr, 2023) – is more straightforward, but hard to do sustainably. It resembles the introduction of neologisms, which can either be new views on existing concepts, or present fully novel concepts. For example, recent conceptual introductions regarding social media are 'enshittification' and 'doomscrolling'. Sometimes this yields a new concept, sometimes a new way to say something we could already think and talk about in less straightforward or apt ways. Both demarcate behaviours and processes that were either previously non-existent or not concisely articulable. A new item expands our collective lexicon that is genuinely distinctive (Cappelen, 2020) – not just some different word for the same concept, like 'stool and chair', which are synonyms, or the trio of 'Venus, morning star and evening star', which are denotations of the same object, but which emphasise different contextual properties. Cappelen (2018, pp. 123–128) writes: "An expression can have cognitive and emotive effects over and beyond (and in some sense independently of) any of its semantic and pragmatic properties". In short, the introduced concept is informationally, cognitively, emotively and referentially novel.[14] Of course, the delineation of concepts is not necessarily *exact*, and conceptual overlap need not be a problem, nor is one of these angles fundamental for the others (Wittgenstein, 2009, §§76-77).

[13] Note however that this example depends on the fact that SPIDER is a strongly referential concept – i.e., tied to bodily entities. We can give ostensive definitions to ground purely referential concepts. For other types of concepts, other forms of improving experiential access are possible.

(3) Lastly, there is *elimination*. Famously, the Churchlands propagated the program of 'eliminativism'. It expected advancements in neuroscience would make folk psychological concepts obsolete and fit for elimination and replacement by scientifically underpinned concepts (Churchland Smith, 1980; Bennett & Hacker, 2003). Steffen Koch's (2023) take on elimination – 'explanatory eliminativism' – is more recent. It holds that revisions can be taken too far, and that existing concepts can prove obsolete by lacking explanatory value. Koch (2023, p. 2141) argues there is a "strong pull" towards elimination because topical limits for revision are overrated. To illustrate, consider the historical examples of Bergson's '*élan vital*' and Stahl's 'phlogiston'. The terms were coined to explain something about life and matter respectively. However, once superseded by scientific advancements, the concepts' expressive powers yielded no additions in explanatory value to theories making use of them. *Ergo*, they have been eliminated from the collective conceptual framework (despite becoming objects of study for the history of philosophy and science).

Now we promptly relate elimination, introduction and amelioration to the three forms of conceptual disruption. *Gaps* can be addressed by the introduction of new concepts or the extension of existing concepts' scope, i.e., gaps can be filled or covered. *Mismatches* can be resolved by providing new delineations of existing concepts – or deliberately allowing for a new blurry overlap – and secondarily by introducing new concepts to cover the area of controversy. Finally, *improvements* to our conceptual network can be made by introducing new concepts to fortify our arsenal, and also by finetuning existing concepts, which is a form of adaptation. But what about elimination? Elimination is a curious case, as it can resolve mismatches *and* lead to improvements by removing dysfunctional concepts. However, it can only address a gap if the eliminated concept is replaced with something else, or if the elimination shows that an apparent gap was unjustified.

[14] As the previous examples tried to show, we have other ways of linking and distinguishing related words, e.g. via synonymy and taxonomy.

Metaphors approximate one conceptual domain through the lens of another conceptual domain. Conceptual engineering modifies a concept's semantic reach (what follows from applying it) and its domain of application (what needs to be present for it to be applicable). Both provide 'access' to our conceptual repertoire in different ways. Regarding drawing a (new) definitional boundary around a concept, like the analysis of conceptual metaphors or exercises in conceptual engineering aim to do, we should be thoughtful that conceptual boundaries are never given, because we "don't know the boundaries because none have been drawn". But this doesn't mean that – following Wittgenstein – we couldn't do so "for a special purpose", making a concept usable for that special purpose (Wittgenstein, 2009, §69). For metaphors and conceptual engineering, this is insightful, as it pre-empts any 'great expectations' like metaphors showing us concepts' 'unchangeable essences', or conceptual engineering supplying us with 'perfectly clear representational devices'. It shows us that the drawing of boundaries defers to the purpose of doing it in that specific way whereas in most metaphorical uses it is the other way around: spontaneous interactions between words lead to 'apparent insights', which are necessarily limited and fuzzy rather than complete and clear in their designations. On the other hand, regarding the controversy surrounding AI-based talk about understanding humans, it shows that if this would be an improvement over non-AI-based framings, improving along that line can only be done *in situ* and *in medias res*, rather than *in vitro* and *ab ovo.*

## 4. The limits and success of the conceptual metaphor view

Let us turn now to a scrutiny of the epistemological limits of applying *CM* to AI-based explanations of the human domain. Appeal to AI concepts to explain human behaviour and mind (e.g., in the stochastic toddler) enacts a conceptual projection. Under *CM* this projection is fundamentally metaphorical; it applies two non-identical conceptual domains to each other. *CM* works as a view to explain the AI-framings because of the latter's structural ubiquity and because of the experimental nature of the AI-projections involved.

In what follows, we unpack *CM*'s applicability to projecting AI concepts on human behaviour, cognition and psychology by showing how it emphasises similarities while suppressing dissimilarities. The similarities *CM* brings to the fore constitute its epistemic success. But the view has clear limits as well that deserve to be explored. Firstly, (4.1.) *CM* risks committing the 'map/territory fallacy': mistaking properties of the projective model for aspects of the target system. Secondly, (4.2.) we give an explanation of AI-based framing of the human domain as a *double metaphor*: built on a metaphorical framing of the computational as cognitive. The technically charged instances of *CM* (layer 1) are prefigured by a modelling move at the foundation of computer science: the Turing machine is underpinned by a metaphorical connection between human calculation and mechanical computation (layer 2).

Conceptual metaphors build epistemic bridges through a set of correspondences between source and target domains, resulting in *emphasis* and *resonance*. The correspondences yield an analogical model for understanding and explaining the target system: a form of explanatory abstraction through analogy. It allows for understanding a complex, opaque system via similarities with a better understood system (Chirimuuta, 2020).[15] This process *abstracts* because for metaphors to resonate and, ultimately, be epistemically successful, highlighted similarities are maximally salient and disanalogies have been minimised. The epistemic tension here is that while metaphors help understand what's otherwise obscure, their imagery guides one's view and favourably spotlights a particular, incomplete perspective of the target system (Möck, 2022). Because of the success of AI-/computer-related conceptualizations, whether metaphorically understood or not, we can rightly ask what other vocabularies would supply better framings. Drawing parallels between AI systems and cognitive processes engenders new insights and could, in principle, foster the development of new hypotheses (Van Rooij et al., 2024; Kompa, 2021). However, AI-related metaphors' abstraction has two clear limits that are important to recognize.

### 4.1. The map-territory fallacy

Over time, the pervasive use of metaphors can lead to the risk of mistaking models (the metaphors) for reality (the actual system). This 'map-territory' fallacy occurs when the analogies (i.e., shared features between domains) are mistakenly regarded as constitutive of the system itself, promoting a conceptual equivalence with the model. As a result, the conceptual metaphor is epistemically stretched too far.

As discussed in previous sections, there has been sustained conceptual borrowing between the brain sciences and AI. As the field of AI rapidly advanced, the cognitive sciences adopted technical and quantifiable constructs from information theory and computer science, framing the brain and mind as computational, information-processing systems — ones that are predictable and quantifiable (Floridi & Nobre, 2024). Over time, these metaphors solidified into standard ways of describing the brain and cognitive phenomena, coming to play a central role in the brain sciences. Computational neuroscientists, for instance, routinely make claims such as: "the olfactory system *computes probability distributions* over possible odours"; "brains *use gradient descent* for learning"; "dopamine neurons *encode reward prediction errors*"; "retinal ganglion cells *are linear filters*" (Colombo & Piccinini, 2024, p. 21, emphasis added). To what extent such claims are metaphorical rather than literal is difficult to determine; we hold that it is, at the very least, uncontroversial to say that the line between them is blurred.

[15] 'Better understood' is relative to a metaphor's target group, e.g. for scientists or lay-people. Given that most modern AI systems are considered to be unexplainable "black-boxes" (Rudin, 2019), this might sound like a contradiction. However, the comparison is usually made at a high level of description (Birhane & McGann, 2024), which allows the easy conceptual transfer between the source and the target system.

As Favela (2022) argues, cognitive psychologists and AI researchers have *literally* meant that brains and minds are computational devices. That is, information-processing systems in the same way that computers are, processing information in the particular ways that computers do. Applied to AI, this implies that both "brains/minds and artificial neural networks manipulate (e.g., firing rates) units (e.g., network nodes) in some sort of systematic way (e.g., encoding and decoding)." In this sense, we argue that the metaphorical conceptual organisation masquerades as 'neutral' scientific investigation or as inference to the best explanation. Like cybernetics, 'information processing' is a wide metaphysical framing, but *performing computations like a computer*, and even more specifically, *computing like a neural network,* have much more concrete inferential consequences. The projection of the ontology of the metaphorical model into the system, as if it weren't a contingent 'net', masks its metaphorical origin.

Putting this form of computationalism aside for now, another problem more relevant to our discussion here is that AI systems are increasingly becoming *working models* of (certain aspects of) human cognition and behaviour (Hassabis et al., 2017; Kriegeskorte & Golan, 2019; Lindsay, 2021; Yamins & DiCarlo, 2016). In short, AI is used as a lens to study these human phenomena, particularly in a way that contains assumptions about *how* the target system processes information. Deep convolutional neural networks are used as models of the visual system – predicting the response of the visual system (Lindsay, 2021) – and large language models are argued to track, under certain conditions, aspects of language acquisition and competence (Millière, 2024). However, the brain-inspired architecture of these models, combined with their success both at performing cognitive tasks and at predicting neural responses, has prompted researchers to argue that AI systems can serve as "plausible simulacra of biological brains" (Hassabis et al., 2017, p. 254) or that deep-learning-style-computations "could plausibly be implemented with biological neurons" (Kriegeskorte & Golan, 2019). The conceptual repertoire developed to describe these models does not stay contained within the modelling context; it migrates into the explanatory vocabulary of cognitive science itself.

To foreshadow part of the argument developed in this section, it is here that the map-territory risk becomes acute: what began as a methodological analogy slides into an ontological claim, projecting the model's particularities back onto the system it was designed to illuminate and bringing with it a whole new conceptual repertoire whose metaphorical nature is difficult to recognize. This might be a reverse instance of what Guest & Van Rooij (2025, p. 4) call the "confusion between statistical versus cognitive models".

The conceptual metaphor, as a *set* of metaphors, constitutes analogical family resemblances between domains. One domain is modelled via another, guiding our understanding. Metaphorical models work when guiding our attention to high-level similarities between systems while obscuring others that are ill-fitting for the model (Möck, 2022), which is the template of abstraction via analogy: "the framing of the investigation of one order of nature by means of its similarity to a system whose order of relations are better known or more readily comprehended by the scientist" (Chirimuuta, 2020, p. 441). The many ways the systems differ are put to the side. Hence, for these metaphorical models to be epistemically fruitful, they have to abstract away the differences between the systems and emphasise their similarities.

Consider the 'brain-computer' metaphor more closely: to comprehend brains and minds as computers requires understanding thinking as a form of computation and, specifically, the kind that digital computers perform — symbol manipulation (see Dupuy, 2009). Conceiving of cognition like this serves as a fundamental guiding commitment of research in AI, cognitive psychology, and neuroscience (Favela, 2022). Such a conceptual framework had epistemic power because of its focus on high-level functions (e.g., input-output processes) while disregarding neurobiological details unshared by computers and brains (Chirimuuta, 2020). This framework excuses the otherwise unjustified separation between a system and its processes, allowing for a focus solely on the *function* of interest. The rest classifies as *just* metabolic support (Chirimuuta, 2024). Therefore, the computational metaphor of the brain is an abstracted model of cognition that focuses on the *function* rather than on the biological substrate supporting it.

The level of abstraction offered by metaphors plays a fundamental role in scientific endeavour (Chirimuuta, 2020). They allow us to prioritize certain features of the studied phenomena that become essential for exploring, hypothesising, explaining, and ultimately understanding some aspects of the target domain (Ibanez et al. 2026; Van Rooij, 2022; Van Rooij et al., 2024). Conceptual metaphors work *because* they abstract away disanalogies and simplify complex concepts into more accessible forms, ultimately emphasising similarities while suppressing differences (Chirimuuta, 2020, 2024). Metaphors are abstracted linguistic models of phenomena.

We observe that metaphorical language is a useful tool for grasping epistemically opaque phenomena. But metaphors' epistemological guidance comes with risk of distracting or manipulating our epistemic access (Möck, 2022). Specifically, the danger lies precisely in metaphors' consistent suppression of differences and amplification of high-level similarities. Recall Kompa (2021) saying that metaphors exploit *contingent* rather than *defining* features. *Overemphasis* of one particular metaphorical model risks gradually eroding our awareness of the distinctions between source and target systems., to the point where "we forget that they ever existed" (Chirimuuta, 2020, p. 441). Thereby the risk of conceptual substitution increases. Conceptual substitution qualifies as gradually occurring conceptual elimination, without justification. This 'conceptual forgetfulness' has been a particularly persistent issue throughout the history of AI (Birhane, 2021). As AI systems increasingly mimic human-like behaviours, conceptual substitution occurs when AI systems are perceived as direct replacements for human cognition (Chirimuuta, 2020). Additionally, our limited understanding of intelligence and overly confident predictions about AI, reinforces the belief that AI concepts explicate essential features of cognition and that we thereby embody such concepts ourselves (Birhane, 2021; Chirimuuta, 2020). We cannot rule out *a priori* that there is an avenue for conceptual engineering, but that avenue needs our *justification*, not *habituation*. We return to this in section 5.

Ignoring differences between AI qua digital, computational machines and biological organisms risks extrapolating features from the AI to human behaviour and psychology. Such an outcome has profound epistemic implications. Among the epistemic risks is that conceptual replacement narrows our perspective of the target system and ends up limiting our understanding of it. Particularly, if we regard organisms only through the lens of the working mechanism of AI algorithms, we might end up overlooking aspects ill-fitted to this comparison, which are fundamental to those systems. We forget the organism is more than its functional replication in the analogy (Chirimuuta, 2020). Moreover, this totalizing view reduces human cognition to sequences of *calculating machines* (Baria & Cross, 2021; Floridi & Nobre, 2024; Fuchs, 2021). To speak with Kompa's criteria for epistemic success: essential properties are obscured, not all inferential patterns can be preserved in the projection, and disanalogies persist.

Returning to the stochastic toddler example, over time we have come to see our cognition and humanity through the lens of AI. The perspective that we are meat-made AI systems isn't bizarre anymore; "i am a stochastic parrot, and so r u". The phrasing implies that humans and AI are fundamentally the same in terms of their core processes: Human and AI language and cognition emerge from statistical computations, extrapolating from patterns learned from the training data supplied by perception, the unsupervised learning of neuronal networks, and the supervised learning of interaction with the world. This reductionist view neglects the unique aspects of human cognition, encouraging the mistaken belief that the similarities are all there is.

However, we shouldn't detract from the possible epistemic benefit of the metaphor as high-level abstraction. There are similarities between human language-use and LLM language production. For example, both humans and AI systems' linguistic activities are grounded in language that is shared, public and historical (Birhane & McGann, 2024). While language learning is differently realized in both systems, in humans and LLMs it is based on exposure, training and reinforcement of pattern responsiveness. In production, both depend on subconscious, autonomous but still context-based responses that incite the reproduction of patterns from a previously modelled informational domain. More could be added to that, but at least it's clear that LLMs provide a system that is to some extent analogous to human language learning and (re)production.

But even though these systems behave in human-like ways, their constitution is fundamentally different (Shanahan, 2024). While their linguistic 'outputs' can be made to converge, human and LLM language 'input' processes strongly diverge, as do the meanings of 'shared' and 'historical' for both systems. Language learning in humans involves much more than just linguistic exposure; a human infant is born into a community of language-users with which it shares a limited world of embodied practices and the acquisition of language involves the interaction with this community and the world they share, whereas a LLM is a disembodied computational entity that, after training on gigantic corpuses of textual data, is able to predict the next word in a sequence of words (tokens) (Shanahan, 2024). As Birhane & McGann (2024, p. 5) aptly emphasise, language is active, embodied and dynamic and involves "voice, text, gestures, body languages, tones, pauses, hesitations, as well as what has been left unsaid". The dimensions of language LLMs can master are only a subset of what human language involves. In Anthony Chemero's (2023) words: "Although humans are quite facile with and can learn quickly from text-based information (just as LLMs do), interacting with text is only one of our ways of knowing about the world around us". Also, even a computationalist like Agüera y Arcas (2025) acknowledges that every AI-based framing by itself cannot account for the breadth of the modelled domain: "A pure reinforcement-learning approach, no matter how fancy the algorithm, cannot account for realistic animal behavior, because real animals—including us—are not optimizing for any one thing (except, maybe, when we're playing championship-level Go)."

Of course, the difference between algorithm talk and LLM talk is that the former entails a similarity in terms of a discretely executable sequence (e.g. of computations), while the latter carries more baggage about the type of computation - how computation is order so to say. The AI domain is a less general tool of description than the computational domain. However, even with regard to saying that cognition 'implements algorithms', we want to be even more careful. The word 'implement' might already go a step too far. We think an algorithm is a mathematical construct that can be utilized to model some empirically delineated target system. So we're okay with the idea that cognition can be algorithmically modelled or approximated, but not with the idea of putting an algorithm into action within cognition, etc. Even though at times LLMs and AI might provide illuminating metaphors for understanding certain aspects of human language-learning-and-production processes, it's crucial to understand that *as metaphor* it's an incomplete representation of human language-use. We should remain wary of confusing the map with the territory.

### 4.2. The double metaphor problem

Turning to the foundations of the concept of computation, we argue that underlying AI-framings is the *double* conceptual metaphor. As Floridi & Nobre (2024) noted, AI research depends on cognitive terms like memory, neuron, stimulus and learning. Similarly, Mitchell (2024a) writes:

> The field of AI has always leaned heavily on metaphors. AI systems are called "agents" that have "knowledge" and "goals"; LLMs are "trained" by receiving "rewards"; "learn" in a "self-supervised" manner by "reading" vast amounts of human-generated text; and "reason" using a method called chain of "thought." These, not to mention the most central terms of the field—*neural* networks, machine *learning*, and artificial *intelligence*—are analogies with human abilities and characteristics that remain quite different from their machine counterparts.

This marks the first layer of the double metaphor: after the initial metaphorical borrowing, technical usages within the AI domain were charged with new technical auras, while being put to new uses, like (un)supervised learning and neural networks.

The second layer is the metaphorical connection underlying the concept of computation itself. A good example comes from Agüera y Arcas, who argues that life – and hence human intelligence – is computational because it is compositional: "A living organism is a compositional structure *par excellence*. It could only be built computationally, through the composition of many functions. And life could only have evolved as the hierarchical composition of those functions—that is, through symbiogenesis." While linking compositionality to symbiogenesis might be a great way of modelling organisms, his framing is strengthened by a reflexive emphasis of its metaphorical roots. We excavate this second layer via a detour through Wittgenstein's thoughts with respect to 'Turing's machines'.

Alan Turing (1937) mechanised the mathematical concept of computation (the human computer). Turing used cognitive terms like 'state of mind', seeing, and remembering, but "Turing machines [do not] demonstrate or possess cognitive abilities; on the contrary, Turing was to stress that 'machine intelligence' only emerges in the shift from 'brute force' to 'learning' programs" (Shanker, 1987, p. 626). His quest revolved only about proving equivalence in terms of expressive power. Turing proposed a mechanical alternative for the human computer, which computes *via* cognitive abilities like seeing, remembering what has been seen and altering what has been remembered according to arithmetical rules, without actually having them. However, Wittgenstein was already aware of the fallacious nature of drawing epistemological conclusions from Turing's success in the mechanisation of calculation through this metaphorical connection. How? According to Shanker (1987, p. 619), Wittgenstein (1980) called Turing's machines "*humans* who calculate", because "Turing [had] actually defined human calculation in mechanical terms so as to license the application of quasi-cognitive terms to the operations of his machines".[16]

Furthermore, Shanker shows that Wittgenstein pre-empted any epistemological conclusions by connecting mechanisation to rule-following. The inequivalence of regularity and rule-following hinges on the logical grammar of rule-following.[17] Turing machines belong to our shared logical fundament, because of the way we *use* recursive effectively calculable functions: they form a logico-grammatical certainty (Shanker, 1987, p. 624; Wittgenstein, 1975). Like certainties, rules fixate the use of concepts and so does the rule of computation that the Turing machine represents: *anything that can be sensibly called an algorithm is executable by a Turing machine*. But this doesn't imply that Turing machines follow meaningless rules because their execution is *mechanical*.

Rather, "mechanically following a rule" is itself problematic, because rule-following is essentially *normative* (Wittgenstein, 2009): it can be evaluated based on the pattern the act of following emulates. The apparent confusion lies in conceiving of algorithms as meaning nothing because of their reducibility to atomic operations. But Turing's machines simply cannot *account for* whether a rule is used for "regulation of their conduct" or whether or not the results that the machines produced were correct (Shanker, 1987, p. 638). Rule-following – like a psychological concept such as reading – is completely independent of any *internal* mechanism (Wittgenstein, 2009, §157), like the 'state of mind' Turing's machine operatively depends on. Sure, rules can be followed 'in a mechanical fashion', but the rule itself should be presentable as ground for doing so.[18] Thus the nexus Turing machine/rule-following is already *metaphorically* rather than metaphysically charged, because the "basic fallacy" of the cognitive framework surrounding Turing machines is "to move from the indisputable simplicity of the sub rules of an algorithm to the conclusion that such procedures are intrinsically mechanical" (Shanker, 1987, p. 641).

[16] Cf. Birhane, 2021.

[17] Supposedly Turing machines 'follow meaningless rules' (i.e., algorithms), and, because they are a mechanical formalisation of human computation, this would imply that human computation is also guided by mechanical rules. The problem with this inference is the idea of a 'meaningless rule'. This problem figures in Wittgenstein's examples of a caveman producing regular sequences of signs and chimpanzees scratching regular figures into the earth: a rule can be constructed to "describe the regularity", but that doesn't mean the observed regularity reduces to following a rule (Shanker, 1987, pp. 620-621).

Connecting this second metaphorical layer to the first layer makes the AI framings potentially *doubly* metaphorical. Making the originally cognitive and behavioural concepts (Turing machine as metaphorical model of human rule-following) explain the human domain again goes full-circle. Others have also noted this full-circle movement, but explain it differently. Barwich & Rodriguez (2024, p. 14) argue the intertwinement of brain and computer

> …unfolded through a dynamic, bidirectional exchange that enhanced our understanding of each system by viewing each through the perspective of the other. Turing's work serves as a prime example of such reciprocal exchange. His insights into computational learning, initially inspired by child development studies […] subsequently informed psychological approaches to childhood education, illustrating a full-circle influence.

Such a "bidirectional exchange" isn't epistemically infertile. However, our investigations lead us to emphasise that whatever terrain is mapped, be it neural, psychological or behavioural, metaphors might help clarify functional aspects of or mechanistic structures in an organism, but only by emphasising specific aspects of a complex phenomenon while *downplaying or ignoring* others.

Metaphors offer a partial and simplified view of the complex realities they aim to represent. Because of these downsides, we shouldn't employ AI metaphors unreflectively – they have epistemic success once we suspend our disbelief in modelling human cognition, behaviour and psychology as stochastic neural networks. So while a computational understanding of cognition doesn't rule out or problematically downplay important facets of cognition, applying LLM-based descriptions metaphorically, does. LLMs are so specific in their workings, modelling of language, and simulation of cognition, that in order for the metaphor to work, we have to suppress a lot of features of human cognition and behaviour. For computation in the broader sense, this isn't necessary.[19] Successful models should surely attract the attention they deserve, but not without the disclaimer that other epistemological conceptual constellations are available and possible and with the upshot in mind that this stance towards the human domain might reveal new insights. (Dennett, 2017). But even though the LLM and AI metaphors are generally limited, one catch remains: for specific cognitive or behavioural concepts the metaphorical semantic charge derived from the conceptual AI domain could still provide avenues for conceptual engineering because of the conceptual metaphor's *descriptive invariance* (Lakoff & Johnson, 1980).

[18] As a sidenote, the rise of LLMs has meant a new wind blowing for the problem of aligning technology to human values. Interestingly, in Pérez-Escobar & Sarikaya (2024) a Wittgensteinian conception of rule-following is drawn upon to inspire how to address the alignment problem in LLMs: how to make them 'act' in accordance with rules like humans can.

## 5. The avenues and challenges for AI-framed conceptual engineering of the human domain

The conceptual metaphor view shows that structural conceptual connections can be made between the human and AI domains. As such, AI advancements could offer a unique, albeit limited, opportunity for conceptual engineering, e.g., refinement of often ambiguous and fuzzy behavioural and psychological concepts. However, caution is advised to prevent oversimplifications or incorporation of a general AI hype into conceptual networks. If any form of *CE* is warranted, this should be because this new conceptual organisation of semantic reach and domain of application yields better results in terms of a population *Y*'s understanding of the phenomenon *X* in question.

Let's first review the opportunities. Subsequently, we explore two challenges that the AI framings of the human domain pose for *CE*.[20] We place addressing these challenges or working through each of the variants of *CE* outside of the scope of this paper and rest content with these pointers for future work.

Note that there are many ways to instantiate *CE*. For purposes of simplicity, we stick to the general view reconstructed in section 3, which consists of amelioration, introduction and elimination. None of us is fully committed to this variant of *CE*, but we and readers with other commitments in terms of *CE* benefit from a fleshing out of what it could look like to ameliorate, introduce and eliminate concepts in a clearly developed case like this.

---

[19] We are grateful to a reviewer for helping us clarify and develop this point.

[20] Deroy (2023, p. 887) discusses three ways in which conceptual approaches to AI and humans can be connected. (1) the *extension view* holds that "naive users extend their category of humans to include AI, even though they do not view AI as central or prototypical in that category"; (2) the *novelty view* is that "naive users have a new category for AI that shares some features with the category of humans but is distinct"; and (3) the *semi-propositional* view, "which proposes that naive users hold a non-fully propositional set of beliefs about AI that do not form a consistent concept, reminiscent of elements of religious beliefs, such as ghosts or spirits". We are inspired by (1) and (2) in what follows.

*Amelioration?* Firstly, it could be that the AI-based projection *extends* human psychological, behavioural and cognitive concepts: widening the domain of application of concepts like thinking, seeing, remembering, feeling to include AI. This *epistemic* amelioration presents anomalies like AI systems as new cases of experiential access.

Alternatively, the full-circle application of AI-concepts could recalibrate the concept of INTELLIGENCE via *semantic* amelioration. The options at hand to do so are changing a concept's pragmatic coordination, logical demarcation, or moral charge. Now, would any of these apply? Firstly, widening the circle of intelligence is an indicator for changing a moral stance towards a being (van Gulick, 2024), which could be a motivating factor for responding to the conceptual muddle like that.[21]

Another motivation that could play a role for moral reasons might move us into the opposite direction: to fix ethical problems, and because of their literal similarity in terms of cognition, extending moral status to nonhuman artificial agents, would be helpful. Secondly, for pragmatic amelioration, a practical social motivator is needed; we *could* come to different agreements about when to speak of intelligence.

We also have the alethic option of repartitioning intelligence's logical space. For example, the ties of intelligence to human forms of life could be loosened. If redefined as 'the capacity to build models', rather than as loose clustering of cognitive capacities ('the intellect'), intelligence's domain of application widens, not only to AI systems, but also to other animals, while partially preserving its core. Alternatively, the standing concept could be tweaked in more subtle ways, such as stripping the criteria related to (likeness to) cognition in order to accommodate new phenomena, i.e., non-cognitive intelligence.

*Expansion?* But maybe the AI-framing doesn't show any problems or imperfections in standing concepts. A possibly less parsimonious way to follow the AI-projection's lead is to coin *new terms for new concepts* by opting for lexical expansion. Not the expression of parallels between human intelligence and AI, but the articulation of a conceptual unifier of artificial and natural intelligence should drive the application. By allowing this to be, say, 'model building' or 'problem solving', a new concept is introduced which expands out of and connects AI and intelligence, but with a different semantic load. However, Cappelen's (2020, pp. 140-146) caveats for illegitimate lexical expansions should be overcome:

1. *preserving lexical effect*: existing terms can incite important moral, legal, cognitive or emotive effects that shouldn't be unnecessarily distorted by new concepts;
2. *topic continuity*: the worry that because words are used over time "*say the same thing*" , keeping the underlying concepts stable is important to facilitate diachronical discourse;
3. *anchoring role*: if the parent concept plays an important role in clustering instances, a lexical expansion with it as starting point might not be such a good idea as it good distort this clustering role; and
4. *role in social ontology*: "changing the meaning of a lexical item might contribute to a change in social reality".

---

[21] But we want to note that Agüera y Arcas (2024) provides a historical contextualisation of this point: "the "incomputably human" narrative", i.e., the idea that "humanness" and "intelligence" cannot be reduced to computation, "is comforting because the Industrial Revolution established a computational hierarchy wherein the soulless, rote, mechanical stuff is done at the lowest tier of the org chart of its time—first by calculating women, then by the machines that replaced them—so that upper class gentlemen-scholars could be freed for the more varied, creative, and intellectual pursuits they deemed worthy of being called "intelligent"."

In short, while it sounds appealing to coin new terms for new forms of expression, it is harder to do than it seems, apart from the fact that for the take-up of such a concept, a need to legitimately embed it into a conceptual practice should be present (Queloz, 2021) such that its reasonably intended also becomes the pragmatic use.

*Elimination?* Given the requirement that some *Y*'s understanding of *X* needs to be unequivocally improved by conceptual engineering, we warn against practising elimination inspired by AI. The danger of opting for conceptual elimination in the guise of conceptual substitution is inspired by the reductionistic character of phrasings such as 'the LLM is a model for human language' or 'the mind as a computer'. While reduction is always tempting, in the *CM*-case of AI it's a one-sided emphasis of the explanatory value of mechanico-computational perspectives at the cost of the messy organismic properties of life and a disregard for the "motley" (Queloz & Cueni, 2021) and "rough ground" (Wittgenstein, 2009, §107) of our conceptual practices. Contrary to spelling out any benefits of conceptual elimination, we advise opting instead for careful amelioration of existing concepts' inferential roles and for the introduction of new concepts. Yet another possibility, as Joe Goegh (2024) argues in another context, is to eliminate the concepts of 'mind' and 'mental' - and, if we follow his line of thought, maybe other psychological concepts as well, because of their disutility and contextual dysfunction, as the conceptual human-AI messy entanglement we are concerned with brings to light.

The main challenge is the *challenge of conceptual ethics* (here we are leaving aside standard problems in conceptual engineering such as samesaying, implementation challenges and whether concepts can even be changed). Not every engineering opportunity is also *desirable*. How to do conceptual ethics is a methodological question of its own, but in the context of AI it is particularly questionable for *whom* it is useful to conceptually engineer concepts from cognition and behaviour so as to fit AI concepts better. Within the AI domain, the ethical question whether it is desirable to widen semantic reach and applicability conditions, or eliminate a concept, should be answered first. For INTELLIGENCE, a conceptual change could affect inclusion into the moral circle, and the extent to which anthropo-exceptionalism is possible by appealing to humans' superior intelligence. Furthermore, some hedged positions on the AI/intelligence relationship are neck-deep into other societal and economic commitments, such as employment or amazement by BigTech and making LLMs and 'AGI' a commercial success. Is there something wrong with conceptual engineering being driven by societal narratives and corporate framings of AI capabilities? Do those pressures and hypes make thinkers and citizens more susceptible to the thought that the AI-human comparison amounts to more than a similarity? Are we in the midst of being *conceptually dominated* by LLM creators into accepting a covert human/AI similarity thesis? Examples where these questions are pressing, are: (legal) personhood for AI; delegation of moral acts to autonomous weapons systems; the ascription of psychological predicates to AI, etc.

Political, cultural or ideological commitments are imported as non-philosophical presuppositions into the debate. An example of this is what Deroy (2023, p. 888) calls "cultural match": different pragmatic usages of "conceptual boundaries" render "official public discourse comprehensible or acceptable to citizens" with disregard for metaphysical boundaries. Alternatively put, legal or cultural recalibration of concepts to accommodate AI can change the conceptual constellations that are tied to specific cultural forms of life of certain social groups and not others, without being necessarily grounded in *actual* differences between humans and AI

But our discussion collides with conceptual ethics going in the opposite direction: the desirability of conceptualizing human concepts in terms of AI, instead of AI in terms of human concepts. An important example on the frontier of the conceptual ethics of AI where a concept from the human domain is in the midst of changing is debate over AGENCY and AI. The extent to which AI systems exhibit autonomous behaviour has affected the way the concept of AGENCY is treated: in many theories, it has become less anthropomorphic and entangled with, e.g., intentionality and 'mindedness' (Floridi, 2025). Also, we read the core of a general worry expressed in Birhane (2021), namely that that the de-anthropomorphization of cognitive, psychological and behavioural concepts qua practice promotes an anthropomorphic conception of AI systems: if brains are like AI, it is easy to think that AI systems are like brains. We need to deal with the question whether it's desirable for us to think and talk about ourselves in terms of AI first. In short: is it desirable to *de-anthropomorphize* the inferential content of human concepts in light of the epistemic success of AI-framings (Cappelen & Dever, 2021; 2025; Mollema, 2026)?

If we blindly follow the lead of the conceptual metaphor towards such conceptual changes and changes in perspective, then, paraphrasing Rosenthal (1982, p. 294), language-users would be normalized towards a "metaphorical naming" that makes political things *invisible.* In other words, there is a danger inherent in solidifying this perspectival shift via an engineering move, as it "predispose[s] us in favor of a specific line of action and it is because metaphors embody proposals that they produce this effect" (Ankersmit, 1993, p. 162). Therefore, conceptual engineering of this domain cannot be done without dealing with the related ethical question of *what* conceptual outcome would serve *whose* purposes. The conceptual metaphor might point in a fruitful direction. But it's up to us to decide whether it's a desirable one, for *us*. In short, the conceptual ethics of AI needs to be done well first.

The second challenge concerns a specific form of reductionism that applying *CE* to AI and human cognition and behaviour can result in. While narrowly defined concepts of cognition may be desirable from a scientific perspective to study its constitutive parts, this view risks flattening the mind's richness to purely mechanistic and reproducible functions (Floridi & Nobre, 2024). The brain isn't *merely* an information-processing or computational apparatus, but a living, plastic, and dynamic system (Favela, 2022; Fuchs, 2021). Anyone framing them in terms of AI concepts should recognize that cognition and the brain are inherently complex systems that might warrant multiple adequate descriptions. We might benefit from a plurality of perspectives deployed in unison in the practices of scientists and laypeople. Ethically, the protection of conceptual pluralism is crucial. Reducing cognition to purely computational or mechanistic terms not only impoverishes our understanding of the mind but also stimulates an unnecessary anthropomorphization of AI systems. This misrepresentation contributes to the ongoing phenomenon of AI hype, where capabilities and performance are overstated (Barrow, 2024; Placani, 2024; Baria & Cross, 2021; Birhane, 2021). From the direction of framing AI in human terms, several contemporary philosophical projects are in the midst of this entanglement, e.g., in the context of moral personhood for AI. Leibo et al. (2025) develop a 'pragmatic' case for AI personhood (see also the debates overviewed in (Gunkel, 2023) and the particular arguments against AI-personhood in (Birhane, et al., 2024) As another example, consider that Bonn (2024) presents a "Moral Turing Test" and claims that AI systems could in principle pass it. The contrary direction, that of framing the human conceptual domain in terms of AI, could also be thought of as contributing to the erosion of conceptual differences between humans and AIs via the structural extension of AI-concepts to human beings or the AI-fication of concepts from the human domain. The challenge of reductionism therefore also points towards the need for doing the conceptual ethics of the conceptual entanglement of human and AI concepts well.

These avenues and challenges open up further debate over the applicability of conceptual engineering to the AI-human nexus.

## 6. Conclusion

In this paper, by articulating and comparing the conceptual metaphor and conceptual engineering approaches to these questions, we showed that the conceptual metaphor projecting the AI domain onto the human domain is severely limited, but bears some epistemic fruits. This approach amounts to the organisation of the human domain in terms of AI-specific computational concepts, which abstracts away from the target domain into an emphasis of shared properties. We argued it faces two problems. First, it risks committing the 'map-territory fallacy': mistaking the model of the target system – built out of computational concepts – for the *actual* target system to be explained. Secondly, by turning to Wittgenstein's reaction to Turing's concept of computation, we explained the appeal of the conceptual metaphor rooted in AI by unmasking it as *doubly metaphorical*: rooted in a metaphorical cognitive connection between the human computer and its mechanical formalisation as Turing machine, and by the full-circle re-application of the concepts *after* being charged with new meanings from AI research. With respect to *CE* the AI-framings might pose avenues for conceptual amelioration or introduction, but faces at least two challenges for AI-inspired conceptual engineering: doing *conceptual ethics* well and avoiding *reductionism*.

Are we all unsupervised learners, stochastically quibbling our way through life? Do some AI concepts provide the 'best game in town' for explaining the human conceptual domain? In Wang's literary example, the temptation of understanding a toddler in terms of AI was in the end immediately quenched by asserting the irreplaceability of human bodily bonds for development. In the article's finale, the mother asserts that "what we say isn't weighted by probability". In other words, by recognizing the irreplaceable aspects of the human, she believes these concepts do not exhaustively make explicit what humans are or do.

The seemingly logical connection between AI-related and organismic concepts is at best an *epistemological perspective;* a *certain* "net" we cast out over the world (Wittgenstein, 1922, 6.342). But alternatives to this organisation are available. On the contrary, Wittgenstein (2009, §132) emphasised that it may "appear as if we saw it as our task to reform language. Such a reform for particular practical purposes, an improvement in our terminology designed to prevent misunderstandings in practice, may well be possible." It is only that, paraphrasing Wittgenstein, the AI-informed epistemological perspective is "an order for a particular purpose, one out of many possible orders, not *the* order."[22] It is plausible that understanding human domains in terms of AI is yet another partially insightful, partially flawed instance of MCO. At its worst, we are only framing the human domain with its limiting case - artificial intelligence: "Seeing a living human being as an automaton is analogous to seeing one figure as a limiting case or variant of another; the cross-pieces of a window as a swastika, for example." (Wittgenstein, 2009, §420). At its best, it's one that prompts us to reflect anew on how the boundaries of our current concepts serve us and how they could be improved.

## Bibliography

Agüera y Arcas, B. (2025). *What is Intelligence?* MIT Press. https://doi.org/10.1162/ANTI.5CZB.

Altman, S. (2022, April 12). *I am a stochastic parrot and so r u* [X]. https://x.com/sama/status/1599471830255177728?lang=es.

Ankersmit, F. R. (1993). Metaphor in Political Theory. In *Knowledge and Language: Volume III: Metaphor and Knowledge* (Ankersmit, F. R. & Mooij, J. J. A. (eds.), pp. 155–202). Springer.

Baria, A. T., & Cross, K. (2021). *The brain is a computer is a brain: Neuroscience's internal debate and the social significance of the Computational Metaphor* (Version 1). arXiv. https://doi.org/10.48550/ARXIV.2107.14042

Barrow, N. (2024). Anthropomorphism and AI hype. *AI and Ethics*, *4*(3), 707–711. https://doi.org/10.1007/s43681-024-00454-1

Barwich, A.-S., & Rodriguez, M. J. (2024). Rage against the what? The machine metaphor in biology. *Biology & Philosophy*, *39*(4), 14. https://doi.org/10.1007/s10539-024-09950-4

Bender, E. M., Gebru, T., McMillan-Major, A., & Shmitchell, S. (2021). On the Dangers of Stochastic Parrots: Can Language Models Be Too Big? 🦜. *Proceedings of the 2021 ACM Conference on Fairness, Accountability, and Transparency*, 610–623. https://doi.org/10.1145/3442188.3445922

Bennett, M., & Hacker, P. M. S. (2003). *Philosophical Foundations of Neuroscience*. Blackwell.

Birhane, A. (2021). The Impossibility of Automating Ambiguity. *Artificial Life*, *27*(1), 44–61. https://doi.org/10.1162/artl_a_00336

[22] This insight is related to what Capraro (2026) rightly notes: "The cultural dominance of the LLM metaphor is not guaranteed: it can be challenged by reviving older metaphors or by introducing new ones. "

Birhane, A., & McGann, M. (2024). Large models of what? Mistaking engineering achievements for human linguistic agency. *Language Sciences*, *106*, 101672. https://doi.org/10.1016/j.langsci.2024.101672

Black, M. (1955). Metaphor. In *Philosophical Perspectives on Metaphor* (Johnson, M. (ed.), pp. 63–82). University of Minnesota Press.

Black, M. (1962). *Models and Metaphors*. Cornell University Press.

Black, M. (1977). More About Metaphor. *Dialectica*, *31*(3/4), 431–457.

Boden, M. A. (1988). *Computer models of mind: Computational approaches in theoretical psychology*. Cambridge University Press.

Bomhof, F. (January 11, 2026). Can submarines swim? *AI Thoughts* https://ai-gedachten.blog/2026/01/11/can-submarines-swim/.

Brette, R. (2022). Brains as Computers: Metaphor, Analogy, Theory or Fact? *Frontiers in Ecology and Evolution*, *10*, 878729. https://doi.org/10.3389/fevo.2022.878729

Cappelen, H. (2018). *Fixing Language: An Essay on Conceptual Engineering*. Oxford University Press.

Cappelen, H. (2020). Conceptual Engineering: The Master Argument. In A. Burgess, H. Cappelen, & D. Plunkett (Eds.), *Conceptual Engineering and Conceptual Ethics* (1st ed., pp. 132–151). Oxford University Press. https://doi.org/10.1093/oso/9780198801856.003.0007

Cappelen, H., & Dever, J. (2021). *Making AI Intelligible: Philosophical Foundations* (1st ed.). Oxford University PressOxford. https://doi.org/10.1093/oso/9780192894724.001.0001

Cappelen, H., & Dever, J. (2025). *Going Whole Hog: A Philosophical Defense of AI Cognition* (Version 1). arXiv. https://doi.org/10.48550/ARXIV.2504.13988

Capraro, V. (2026). LLMorphism: When humans come to see themselves as language models. https://doi.org/10.48550/arXiv.2605.05419

Chemero, A. (2023). LLMs differ from human cognition because they are not embodied. *Nature Human Behaviour*, *7*(11), 1828–1829. https://doi.org/10.1038/s41562-023-01723-5

Chirimuuta, M. (2020). The Reflex Machine and the Cybernetic Brain: The Critique of Abstraction and its Application to Computationalism. *Perspectives on Science*, *28*(3), 421–457. https://doi.org/10.1162/posc_a_00346

Chirimuuta, M. (2021). Your Brain Is Like a Computer: Function, Analogy, Simplification. In F. Calzavarini & M. Viola (Eds.), *Neural Mechanisms* (Vol. 17, pp. 235–261). Springer International Publishing. https://doi.org/10.1007/978-3-030-54092-0_11

Chirimuuta, M. (2024). *The Brain Abstracted: Simplification in the History and Philosophy of Neuroscience*. The MIT Press.

Churchland Smith, P. (1980). A perspective on mind-brain research. *Journal of Philosophy*, *77*, 185–207.

Cobb, M. (2020). *The Idea of the Brain: The Past and Future of Neuroscience*. Profile Books.

Colombo, M., & Piccinini, G. (2024). *The Computational Theory of Mind* (1st ed.). Cambridge University Press. https://doi.org/10.1017/9781009183734

Dennett, D. C. (2017). *From Bacteria to Bach and Back: The Evolution of Minds*. Allen Lane.

Deroy, O. (2023). The Ethics of Terminology: Can We Use Human Terms to Describe AI? *Topoi*, *42*(3), 881–889. https://doi.org/10.1007/s11245-023-09934-1

Dupuy, J.-P. (2009). *On the origins of cognitive science: The mechanization of the mind* (M. B. DeBevoise, Trans.). MIT Press.

Favela, L. (2022). Complexity: Understanding brains and minds on their own terms. *Preprint*. https://www.researchgate.net/publication/358402375_Complexity_Understanding_brains_and_minds_on_their_own_terms

Felin, T., & Holweg, M. (2024). Theory Is All You Need: AI, Human Cognition, and Decision Making. *SSRN Electronic Journal*. https://doi.org/10.2139/ssrn.4737265

Floridi, L., & Nobre, A. C. (2024). Anthropomorphising Machines and Computerising Minds: The Crosswiring of Languages between Artificial Intelligence and Brain & Cognitive Sciences. *Minds and Machines*, *34*(1), 5. https://doi.org/10.1007/s11023-024-09670-4

Floridi, L. (2025). AI as Agency without Intelligence: On Artificial Intelligence as a New Form of Artificial Agency and the Multiple Realisability of Agency Thesis. *Philosophy & Technology* 38, 30. https://doi.org/10.1007/s13347-025-00858-9

Fuchs, T. (2021). Human and Artificial Intelligence: A Clarification. In T. Fuchs, *In Defence of the Human Being* (1st ed., pp. 13–48). Oxford University PressOxford. https://doi.org/10.1093/oso/9780192898197.003.0002

Gough, J. (2024). 'Mind' and 'mental': extended, pluralistic, eliminated. *Synthese* 204, 146. https://doi.org/10.1007/s11229-024-04788-5

Guest, O., & van Rooij, I. (2025). Critical Artificial Intelligence Literacy for Psychologists. https://doi.org/10.31234/osf.io/dkrgj_v1

Haslanger, S. (2020). Going On, Not in the Same Way. In A. Burgess, H. Cappelen, & D. Plunkett (Eds.), *Conceptual Engineering and Conceptual Ethics* (1st ed., pp. 230–260). Oxford University PressOxford. https://doi.org/10.1093/oso/9780198801856.003.0012

Hassabis, D., Kumaran, D., Summerfield, C., & Botvinick, M. (2017). Neuroscience-inspired artificial intelligence. Neuron, 95(2), 245–258. https://doi.org/10.1016/j.neuron.2017.06.011

Hopster, J., Brey, P., Klenk, M., Löhr, G., Marchiori, S., Lundgren, B., & Scharp, K. (2023). 6. Conceptual Disruption and the Ethics of Technology. In I. Van De Poel, L. E. Frank, J. Hermann, J. Hopster, D. Lenzi, S. Nyholm, B. Taebi, & E. Ziliotti (Eds.), *Ethics of Socially Disruptive Technologies* (1st ed., pp. 141–162). Open Book Publishers. https://doi.org/10.11647/obp.0366.06

Hopster, J., & Löhr, G. (2023). Conceptual Engineering and Philosophy of Technology: Amelioration or Adaptation? *Philosophy & Technology*, *36*(4), 70. https://doi.org/10.1007/s13347-023-00670-3

Ibanez, A., Roth, N., Colverson, A., Bailey, C., Miller, B., Durón-Reyes, D. E., Johnson, N., Castaner, O., Sacco, P. L., Cotter, E., & Melloni, L. (2026). Music as a scientific metaphor for mind and brain. *Neuroscience & Biobehavioral Reviews*, 185, 106643. https://doi.org/10.1016/j.neubiorev.2026.106643

Koch, S. (2023). Why Conceptual Engineers Should Not Worry About Topics. *Erkenntnis*, *88*(5), 2123–2143. https://doi.org/10.1007/s10670-021-00446-1

Kompa, N. (2021). Insight by Metaphor – The Epistemic Role of Metaphor in Science. In A. Heydenreich & K. Mecke (Eds.), *Physics and Literature* (pp. 23–48). De Gruyter. https://doi.org/10.1515/9783110481112-002

Kövecses, Z. (2008). Conceptual metaphor theory: Some criticisms and alternative proposals. *Annual Review of Cognitive Linguistics*, *6*, 168–184. https://doi.org/10.1075/arcl.6.08kov

Kövecses, Z. (2017). Conceptual Metaphor Theory: Some New Proposals. *LaMiCuS*, *1*(1), 16–32.

Kriegeskorte, N., & Golan, T. (2019). Neural network models and deep learning. *Current biology: CB*, 29(7), R231–R236. https://doi.org/10.1016/j.cub.2019.02.034

Lakoff, G. (1993). The contemporary theory of metaphor. In A. Ortony (Ed.), *Metaphor and Thought* (2nd ed., pp. 202–251). Cambridge University Press. https://doi.org/10.1017/CBO9781139173865.013

Lakoff, G., & Johnson, M. (1981). Conceptual Metaphor in Everyday Language. In *Philosophical Perspectives on Metaphor* (Johnson, M. (ed.), pp. 286–325). University of Minnesota Press.

Lakoff, G., & Johnson, M. (2003). *Metaphors We Live By*. University of Chicago Press.

Lewontin, R. C. (1996). Evolution as Engineering. In J. Collado-Vides, B. Magasanik, & T. F. Smith (Eds.), *Integrative Approaches to Molecular Biology* (pp. 1–10). The MIT Press. https://doi.org/10.7551/mitpress/3824.003.0002

Lindsay G. W. (2021). Convolutional Neural Networks as a Model of the Visual System: Past, Present, and Future. *Journal of cognitive neuroscience*, *33*(10), 2017–2031. https://doi.org/10.1162/jocn_a_01544

Löhr, G. (2024). If conceptual engineering is a new method in the ethics of AI, what method is it exactly? *AI and Ethics*, *4*(2), 575–585. https://doi.org/10.1007/s43681-023-00295-4

McCarthy, J., Minsky, M. L., Rochester, N., & Shannon, C. E. (2006). A Proposal for the Dartmouth Summer Research Project on Artificial Intelligence, August 31, 1955. *AI Magazine*, *27*(4), 12. https://doi.org/10.1609/aimag.v27i4.1904

Miller, E. F. (1979). Metaphor and Political Knowledge. *The American Political Science Review*, *73*(1), 155–170.

Millière, R. (2024). *Language models as models of language*. arXiv. https://arxiv.org/abs/2408.07144

Mitchell, M. (2024a). The metaphors of artificial intelligence. *Science*, *386*(6723), eadt6140. https://doi.org/10.1126/science.adt6140

Mitchell, M. (2024b). The Turing Test and our shifting conceptions of intelligence. *Science*, *385*(6710), eadq9356. https://doi.org/10.1126/science.adq9356

Möck, L. A. (2022). Prediction Promises: Towards a Metaphorology of Artificial Intelligence. *Journal of Aesthetics and Phenomenology*, *9*(2), 119–139. https://doi.org/10.1080/20539320.2022.2143654

Mollema, W. J. T. (2024a). *On certainty*, Left Wittgensteinianism and conceptual change. *Theoria*, theo.12558. https://doi.org/10.1111/theo.12558

Mollema, W. J. T. (2024b). Social AI and the Equation of Wittgenstein's Language User with Calvino's Literature Machine. *International Review of Literary Studies*, *6*(1), 39–55.

Mollema, W. J. T. (2026). On Large Language Models, De-anthropomorphized Narrative Production, and Deflationary Understanding: A Reply to Rodrigues. *Philosophy & Technology*, *39*(1), 10. https://doi.org/10.1007/s13347-025-01019-8

Neisser, U. (1967). *Cognitive psychology*. Appleton-Century-Crofts.

Nicholson, D. J. (2013). Organisms≠Machines. *Studies in History and Philosophy of Science Part C: Studies in History and Philosophy of Biological and Biomedical Sciences*, *44*(4), 669–678. https://doi.org/10.1016/j.shpsc.2013.05.014

Odell, S. J. (2001). Wittgenstein and Computationalism. *From the ALWS Archives: A Selection of Papers from the International Wittgenstein Symposia in Kirchberg Am Wechsel*. https://wab.uib.no/ojs/index.php/agora-alws/article/view/2412

Pérez-Escobar, J. A., & Sarikaya, D. (2024). Philosophical Investigations into AI Alignment: A Wittgensteinian Framework. *Philosophy & Technology*, *37*(3), 80. https://doi.org/10.1007/s13347-024-00761-9

Piccinini, G. (2006). Computational explanation in neuroscience. *Synthese*, *153*(3), 343–353. https://doi.org/10.1007/s11229-006-9096-y

Piccinini, G. (2009). Computationalism in the Philosophy of Mind. *Philosophy Compass*, *4*(3), 515–532. https://doi.org/10.1111/j.1747-9991.2009.00215.

Placani, A. (2024). Anthropomorphism in AI: Hype and fallacy. *AI and Ethics*, *4*(3), 691–698. https://doi.org/10.1007/s43681-024-00419-4

Queloz, M. (2021). *The Practical Origins of Ideas: Genealogy as Conceptual Reverse-Engineering*. Oxford University Press.

Queloz, M. (2025). *The Ethics of Conceptualization: Tailoring Thought and Language to Need*. Oxford University Press.

Queloz, M., & Cueni, D. (2021). Left Wittgensteinianism. *European Journal of Philosophy*, *29*(4), 758–777. https://doi.org/10.1111/ejop.12603

Rivadulla, A. (2006). Metáforas y modelos en ciencia y filosofía. *Revista de Filosofía (Madrid)*, *31*(2), 189–202.

Rosenthal, D. C. (1982). Metaphors, Models, and Analogies in Social Science and Public Policy. *Political Behavior*, *4*(3), 283–301.

Rudin, C. (2019). Stop explaining black box machine learning models for high stakes decisions and use interpretable models instead. *Nature Machine Intelligence*, *1*(5), 206–215. https://doi.org/10.1038/s42256-019-0048-x

Shagrir, O. (2006). Why we view the brain as a computer. *Synthese*, *153*(3), 393–416. https://doi.org/10.1007/s11229-006-9099-8

Shanahan, M. (2024). *Simulacra as conscious exotica*. arXiv. https://arxiv.org/abs/2402.12422

Shanker, S. G. (1987). Wittgenstein versus Turing on the Nature of Church's Thesis. *Notre Dame Journal of Formal Logic*, *28*(4), 615–649.

Summerfield, C. (2023). Natural general intelligence: How understanding the brain can help us build AI. Oxford University Press. https://doi.org/10.1093/oso/9780192843883.001.0001

Taylor, C., & Dewsbury, B. M. (2018). On the Problem and Promise of Metaphor Use in Science and Science Communication. *Journal of Microbiology & Biology Education*, *19*(1), 19.1.40. https://doi.org/10.1128/jmbe.v19i1.1538

Turing, A. M. (1937). On Computable Numbers, with an Application to the Entscheidungsproblem. *Proceedings of the London Mathematical Society*, *s2-42*(1), 230–265. https://doi.org/10.1112/plms/s2-42.1.230

van Gulick, R. (2024). Consciousness. In *The Stanford Encyclopedia of Philosophy* (Winter 2022 Edition, Zalta, E. N. & Nodelman, U. (eds.).). https://plato.stanford.edu/archives/win2022/entries/consciousness/

Van Rooij, I. (2022). Psychological models and their distractors. *Nature Reviews Psychology*, *1*(3), 127–128. https://doi.org/10.1038/s44159-022-00031-5

Van Rooij, I., Guest, O., Adolfi, F., De Haan, R., Kolokolova, A., & Rich, P. (2024). Reclaiming AI as a Theoretical Tool for Cognitive Science. *Computational Brain & Behavior*. https://doi.org/10.1007/s42113-024-00217-5

Veluwenkamp, H., Hopster, J., Köhler, S., & Löhr, G. (2024). Socially Disruptive Technologies and Conceptual Engineering. *Ethics and Information Technology*, *26*(4), 65. https://doi.org/10.1007/s10676-024-09804-3

Wang, A. (2023, November 13). Is My Toddler A Stochastic Parrot? *The New Yorker*. https://www.newyorker.com/humor/sketchbook/is-my-toddler-a-stochastic-parrot

Wittgenstein, L. (1922). *Tractatus Logico-Philosophicus* (Trans. Ogden, C. K.). Kegan Paul, Trench, Trubner & Co.

Wittgenstein, L. (1975). *On Certainty* (Anscombe, G. E. M. & von Wright, G. H. (eds.). Trans. Paul, D.). Blackwell.

Wittgenstein, L. (1980). *Remarks on the Philosophy of Psychology, Volume I* (Trans. Anscombe, G. E. M.. Anscombe, G. E. M. and von Wright, G.H. (eds.)). Basil Blackwell.

Wittgenstein, L. (2009). *Philosophical Investigations* (Trans. Anscombe, G. E. M., Hacker, P. M. S. & Schulte, J.). Wiley-Blackwell.

Yamins, D., DiCarlo, J. Using goal-driven deep learning models to understand sensory cortex. *Nat Neurosci* 19, 356–365 (2016). https://doi.org/10.1038/nn.4244